\makeatletter\def\graphicscache@inhibit{true}\makeatother
\documentclass[a4paper,twoside]{article}

\usepackage{graphics} %
\usepackage{amsmath} %
\usepackage{amssymb}  %

\usepackage{graphicscache}
\usepackage{multirow}

\usepackage{booktabs}
\usepackage{threeparttable}

\usepackage[final,nomargin,inline]{fixme}
\fxsetup{theme=color}

\usepackage[hidelinks]{hyperref}
\usepackage[style=ieee,hyperref,natbib=true,backend=bibtex,firstinits,doi=false,%
     mincitenames=1,maxcitenames=2,maxbibnames=99,sorting=none,terseinits=false,hyperref=true]{biblatex}
\bibliography{references.bib}
\renewbibmacro*{bbx:savehash}{}%
\defbibheading{bibliography}[\bibname]{\section*{References}}

\usepackage[capitalize]{cleveref}

\usepackage{tikz}
\usetikzlibrary{calc}

\usepackage{multicol}
\usepackage{pslatex}
\usepackage{epsfig}
\usepackage{SCITEPRESS}     %

\renewcommand\orcidAuthor[1]{
\hspace*{-1mm}\includegraphics[width=0.3cm]{orcid.pdf}\thanks{\protect\includegraphics[width=0.3cm]{orcid.pdf}~https://orcid.org/#1}\hspace*{-1mm}
}

\begin{document}

\title{ConvPoseCNN: Dense Convolutional 6D Object Pose Estimation}

\author{
\authorname{Catherine Capellen \sup{1}%
,  Max Schwarz\sup{1} \orcidAuthor{0000-0002-9942-6604}
and Sven Behnke\sup{1} \orcidAuthor{0000-0002-5040-7525}}
\affiliation{\sup{1}Autonomous Intelligent Systems group of University of Bonn, Germany}
\email{max.schwarz@ais.uni-bonn.de}
}

\keywords{Pose Estimation, Dense Prediction, Deep Learning}

\abstract{%
6D object pose estimation is a prerequisite for many applications.
In recent years, monocular pose estimation has attracted much research
interest because it does not need depth measurements.
In this work, we introduce ConvPoseCNN, a fully convolutional architecture that avoids cutting out individual objects.
Instead we propose pixel-wise, dense prediction of both translation and orientation components of the object pose,
where the dense orientation is represented in Quaternion form.
We present different approaches for aggregation of the dense orientation predictions,
including averaging and clustering schemes.
We evaluate ConvPoseCNN on the challenging YCB-Video Dataset, where we show that
the approach has far fewer parameters and trains faster than comparable methods
without sacrificing accuracy.
Furthermore, our results indicate that the dense orientation prediction implicitly
learns to attend to trustworthy, occlusion-free, and feature-rich object regions.
}

\onecolumn
\maketitle \normalsize \setcounter{footnote}{0} \vfill

\section{Introduction}

Given images of known, rigid objects, 6D object pose estimation describes the problem of determining the identity of the objects, their position and their orientation. 
Recent research focuses on increasingly difficult datasets with multiple objects per image, cluttered environments, and partially occluded objects.
Symmetric objects pose a particular challenge for orientation estimation, because multiple solutions or manifolds of solutions exist.
While the pose problem mainly receives attention from the computer vision community,
in recent years there have been multiple robotics competitions involving
6D pose estimation as a key component, for example the Amazon Picking Challenge
of 2015 and 2016 and the Amazon Robotics Challenge of 2017,
where robots had to pick objects from highly cluttered bins.
The pose estimation problem is also highly relevant in human-designed, less structured environments,
e.g. as encountered in the RoboCup@Home competition~\cite{iocchi2015robocup}, where
robots have to operate within home environments.
\section{Related Work}

For a long time, feature-based and template-based methods were popular for 6D
object pose estimation \citep{lowe2004distinctive, wagner2008pose, hinterstoisser2012gradient, hinterstoisser2012model}.
However, feature-based methods rely on distinguishable features and perform badly for texture-poor objects. Template-based methods do not work well if objects are partially occluded.
With deep learning methods showing success for different image-related problem settings, models inspired or extending these have been used increasingly. Many methods use established architectures to solve sub-problems, as for example semantic segmentation or instance segmentation. Apart from that, most recent methods use deep learning for their complete pipeline.
We divide these methods into two groups:
Direct pose regression methods \citep{deep6dpose, xiang2017posecnn} and methods that
predict 2D-3D object correspondences and then solve the PnP problem to recover the 6D pose.
The latter can be further divided into methods that predict dense, pixel-wise correspondences \citep{brachmann2014learning, brachmann2016uncertainty, krull2015learning} and, more recently, methods that estimate the 2D coordinates of selected keypoints, usually the 3D object bounding box corners \citep{heatmaps, BB8,  SSS6D, dope}.

\citet{heatmaps} predict the projection of the 3D bounding box as a heat map. To achieve robustness to occlusion, they predict the heat map independently for small object patches before adding them together. The maximum is selected as the corner position. If patches are ambiguous, the training technique implicitly results in an ambiguous heat map prediction. This method also uses Feature Mapping \citep{featuremapping}, a technique to bridge the domain-gap between synthetic and real training data.

\begin{figure}[t]
\includegraphics[width=0.48\textwidth]{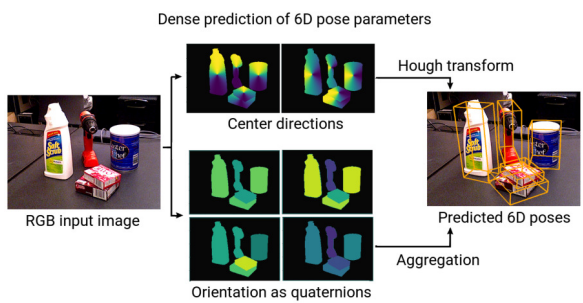}
\caption{Dense Prediction of 6D pose parameters inside ConvPoseCNN. The dense
 predictions are aggregated on the object level to form 6D pose outputs.}
 \label{fig:teaser}
\end{figure}

\begin{figure*}
\centering
\includegraphics[width=0.8\textwidth]{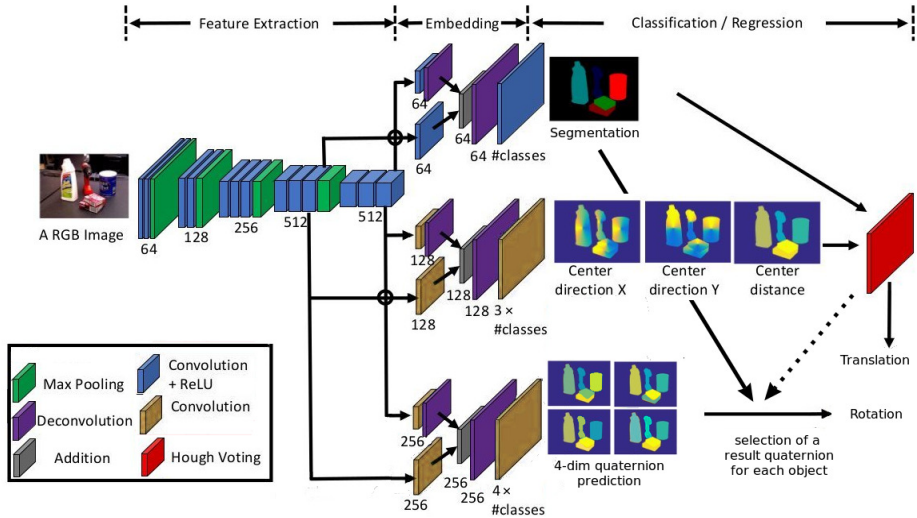}
\caption{Our ConvPoseCNN architecture for convolutional pose estimation.
During aggregation, candidate quaternions are selected according to
the semantic segmentation results or according to Hough inlier information.
Figure adapted from \citep{xiang2017posecnn}.}
\label{pic:convolutionalnetwork}
\end{figure*}

We note that newer approaches increasingly focus on the monocular pose estimation
problem without depth information (\citep{brachmann2016uncertainty, deep6dpose, jafari2017ipose, BB8, heatmaps, SSS6D, dope, xiang2017posecnn}).
In addition to predicting the pose from RGB or RGB-D data, there are several refinement techniques for pose improvement after the initial estimation. \Citet{li2018deepim} introduce a render-and-compare technique that improves the estimation only using the original RGB input.
If depth is available, ICP registration can be used to refine poses.

As a representative of the direct regression method, we discuss PoseCNN \citep{xiang2017posecnn}
in more detail. It delivered state-of-the-art performance on the occluded LINEMOD dataset and introduced a more challenging dataset, the YCB-Video Dataset.
PoseCNN decouples the problem of pose estimation into estimating the translation and orientation separately.
A pretrained VGG16 backbone is used for feature extraction. The features are processed in three different branches:
Two fully convolutional branches estimate a semantic segmentation, center directions, and the depth for every pixel of the image. The third branch consists of a RoI pooling and a fully-connected architecture which regresses to a quaternion describing the rotation for each region of interest.

RoI pooling -- i.e. cutting out and size normalizing an object hypothesis -- was originally developed for the object detection problem \citep{girshick2015fast},
where it is used to extract an object-centered and size-normalized view of
the extracted CNN features. The following classification network, usually consisting
of a few convolutional and fully-connected layers, then directly computes class scores
for the extracted region.
As RoI pooling focusses on individual object hypotheses, it looses contextual information, which might be important in 
cluttered scenes where objects are densely packed and occlude each other. 
RoI pooling requires random access to the source feature map for cutting out and interpolating features.
Such random access patterns are expensive to implement in hardware circuits and
have no equivalent in the visual cortex \citep{kandel2000principles}.
Additionally, RoI pooling is often followed by
fully connected layers, which drive up parameter count and inference/training time.

Following the initial breakthroughs using RoI pooling, simpler architectures
for object detection have been proposed which compute the class scores
in a fully convolutional way \citep{redmon2016you}. An important insight here
is that a CNN is essentially equivalent to a sliding-window operator, i.e.
fully-convolutional classification is equivalent to RoI-pooled classification with a fixed region
size.
While the in-built size-invariance of RoI pooling is lost, fully-convolutional architectures
typically outperform RoI-based ones in terms of model size and training/inference speed.
With a suitably chosen loss function that addresses the inherent example imbalances
during training \citep{lin2017focal}, fully-convolutional architectures 
reach state-of-the-art accuracy in object detection.

Following this idea, we developed a fully-convolutional architecture evolved
from PoseCNN, that replaces the RoI pooling-based orientation estimation of PoseCNN
with a fully-convolutional, pixel-wise quaternion orientation prediction (see \cref{fig:teaser}).
Recently, \citet{peng2019pvnet} also removed the RoI-pooled orientation prediction
branch, but with a different method: Here, 2D directions to a fixed number of
\textit{keypoints} are densely predicted. Each keypoint is found using a
separate Hough transform and the pose is then estimated using a PnP solver
utilizing the known keypoint correspondences.
In contrast, our method retains the direct orientation regression branch, which
may be interesting in resource-constrained scenarios, where the added
overhead of additional Hough transforms and PnP solving is undesirable.

Our proposed changes unify the architecture and make it more parallel: PoseCNN first predicts the translation and the regions of interest (RoI) and then, sequentially for each RoI estimates object orientation. Our architecture can perform the rotation estimation for multiple objects in parallel, independent from the translation estimation.
We investigated different averaging and clustering schemes for obtaining a final orientation from our pixel-wise estimation. We compare the results of our architecture to PoseCNN on the YCB-Video Dataset \citep{xiang2017posecnn}.
We show that our fully-convolutional architecture with pixel-wise prediction achieves precise results while using far less parameters. The simpler architecture also results in shorter training times.

In summary, our contributions include:
\begin{enumerate}
 \item A conceptually simple, small, and fast-to-train architecture for dense orientation estimation,
   whose prediction is easily interpretable due to its dense nature,
 \item a comparison of different orientation aggregation techniques, and
 \item a thorough evaluation and ablation study of the different design choices on the
   challenging YCB-Video dataset.
\end{enumerate}

\section{Method}

We propose an architecture derived from PoseCNN \citep{xiang2017posecnn},
which predicts, starting from RGB images, 6D poses for each object in the image.
The network starts with the convolutional backbone of VGG16 \citep{simonyan2014very} that  extracts features.
These are subsequently processed in three branches: The fully-convolutional segmentation branch that predicts a pixel-wise semantic segmentation, the fully-convolutional vertex branch, which predicts a pixel-wise estimation of the center direction and center depth, and the quaternion estimation branch.
The segmentation and vertex branch results are combined to vote for object centers in a Hough transform layer. 
The Hough layer also predicts bounding boxes for the detected objects. 
PoseCNN then uses these bounding boxes to crop and pool the extracted features which are then fed into a fully-connected neural network architecture. This fully-connected part predicts an orientation quaternion for each bounding box.

Our architecture, shown in \cref{pic:convolutionalnetwork}, replaces the quaternion estimation branch of PoseCNN with a fully-convolutional architecture, similar to the segmentation and vertex prediction branch. It predicts quaternions pixel-wise. We call it ConvPoseCNN (short for convolutional PoseCNN).
Similarly to PoseCNN, quaternions are regressed directly using a linear output layer.
The added layers have the same architectural parameters as in the segmentation branch (filter size 3$\times$3)
and are thus quite light-weight.

While densely predicting orientations at the pixel level might seem
counter-intuitive, since orientation estimation typically needs long-range information
from distant pixels, we argue that due to the total depth of the convolutional
network and the involved pooling operations the receptive field for a single
output pixel covers large parts of the image and thus allows long-range
information to be considered during orientation prediction.

\subsection{Aggregation of Dense Orientation Predictions}
\label{chap:averagingquats}

We estimate quaternions pixel-wise and use the predicted segmentation to identify which quaternions belong to which object. If multiple instances of one object can occur, one could use the Hough inliers instead of the segmentation.
Before the aggregation of the selected quaternions to a final orientation estimate,
we ensure that each predicted quaternion $q$ corresponds to a rotation by
scaling it to unit norm.
However, we found that the norm $w=||q||$ prior to scaling is of interest
for aggregation: In feature-rich regions, where there is more evidence for
the orientation prediction, it tends to be higher (see \cref{subsec:translationresults}).
We investigated averaging and clustering techniques for aggregation, optionally
weighted by $w$.

For averaging the predictions we use the weighted quaternion average as defined by \citet{markley2007quaternion}.
Here, the average $\bar{q}$ of quaternion samples $q_1,...,q_n$ with weights $w_1,...,w_n$ is defined
using the corresponding rotation matrices $R(q_1),...,R(q_n)$:
\begin{equation}
 \bar{q} = \text{arg} \min_{q \in \mathbb{S}^3} \sum_{i=1}^n w_i ||R(q) - R(q_i)||^2_F,
\end{equation}
where $\mathbb{S}^3$ is the unit 3-sphere and $||\cdot||_F$ is the Frobenius norm.
This definition avoids any problems arising from the antipodal symmetry of
the quaternion representation. The exact solution to the optimization problem
can be found by solving an eigenvalue problem \citep{markley2007quaternion}.

For the alternative clustering aggregation, we follow a weighted RANSAC scheme:
 For quaternions $Q=\{q_1, ..., q_n\}$ and their weights $w_1, ..., w_n$  associated with one object this algorithm repeatedly chooses a random quaternion $\hat{q} \in Q$ with a probability proportional to its weight and then
determines the inlier set $\bar{Q}=\{q \in Q|d(q,\hat{q}) < t\}$, where $d(\cdot,\cdot)$ is the angular distance.
Finally, the $\hat{q}$ with largest $\sum_{q_i \in \bar{Q}} w_i$ is selected as the result quaternion.

The possibility of weighting the individual samples is highly useful in this
context, since we expect that parts of the object are more important for
determining the correct orientation than others (e.g. the handle of a cup).
In our architecture, sources of such pixel-wise weight information can be the
segmentation branch with the class confidence scores, as well as the predicted
quaternion norms $||q_1||,...,||q_n||$ before normalization.

\subsection{Losses and Training}
\label{sec:losses}

For training the orientation branch, \citet{xiang2017posecnn} propose the ShapeMatch loss.
This loss calculates a distance measure between point clouds of the object
rotated by quaternions $\tilde{q}$ and $q$:
\begin{equation}
 \text{SMLoss}(\tilde{q},q) = \begin{cases}
                                       \text{SLoss}(\tilde{q},q), & \text{if object is symmetric,} \\
                                       \text{PLoss}(\tilde{q},q), & \text{otherwise.}
                                      \end{cases},
\end{equation}

Given a set of 3D points $\mathbb{M}$, where 
 m = $|\mathbb{M}|$ and $R(q)$ and $R(\tilde{q})$ are the rotation matrices corresponding to ground truth and estimated quaternion, respectively, and Ploss and Sloss are defined in \citep{xiang2017posecnn} as follows:
\begin{alignat}{2}
\text{PLoss}(\tilde{q},q) &= \frac{1}{2m} \sum_{x\in \mathbb{M}} || R(\tilde{q})x - R(q)x||^2, \\
\text{SLoss}(\tilde{q},q) &= \frac{1}{2m} \sum_{x_1\in \mathbb{M}}\min_{x_2\in\mathbb{M}} || R(\tilde{q})x_1 - R(q)x_2||^2.
\end{alignat}
Similar to the ICP objective, SLoss does not penalize rotations of symmetric
objects that lead to equivalent shapes.

In our case, ConvPoseCNN outputs a dense, pixel-wise orientation prediction.
Computing the SMLoss pixel-wise is computationally prohibitive.
First aggregating the dense predictions and then calculating the orientation loss makes it possible to
train with SMLoss. In this setting, we use a naive average, the normalized sum of all quaternions,
to facilitate backpropagation through the aggregation step.
As a more efficient alternative we experiment with pixel-wise L2 or Qloss \citep{billings2018silhonet}
loss functions, that are evaluated for the pixels indicated by the ground-truth segmentation.
Qloss is designed to handle the quaternion symmetry. For two quaternions $\bar{q}$ and $q$ it is defined as:
\begin{equation}
\text{Qloss}(\bar{q}, q) = \log(\epsilon + 1 - |\bar{q}\cdot q|),
\end{equation}
where $\epsilon$ is introduced for stability.

The final loss function used during training is, similarly to PoseCNN, a linear
combination of segmentation ($L_{\text{seg}}$), translation ($L_{\text{trans}}$), and orientation loss ($L_{\text{rot}}$):
\begin{equation}
L = \alpha_{\text{seg}}L_{\text{seg}} + \alpha_{\text{trans}}L_{\text{trans}} + \alpha_{\text{rot}}L_{\text{rot}}.
\end{equation}

\section{Evaluation}

\subsection{Datasets}

We perform our experiments on the challenging YCB-Video Dataset~\citep{xiang2017posecnn}.
The dataset contains 133,936 images extracted from 92 videos, showing 21 rigid objects. For each object the dataset contains a point model with 2620 points each and a mesh file.
Additionally the dataset contains 80.000 synthetic images. The synthetic images are not physically realistic. Randomly selected images from SUN2012 \citep{xiao2010sun} and ObjectNet3D \citep{xiang2016objectnet3d} are used as backgrounds for the synthetic frames.

When creating the dataset only the first frame of each video was annotated manually and the rest of the frames were inferred using RGB-D SLAM techniques. Therefore, the annotations are sometimes less precise.

The images contain multiple relevant objects in each image, as well as occasionally uninteresting objects and distracting background. Each object appears at most once in each image. The dataset includes symmetric and texture-poor objects, which are especially challenging.

\subsection{Evaluation Metrics}
We evaluate our method under the AUC P and AUC S metrics as defined for PoseCNN~\citep{xiang2017posecnn}. For each model we report the total area under the curve for all objects in the test set. The AUC P variant is based on a point-wise
distance metric which does not consider symmetry effects (also called ADD). In contrast, AUC S is based on an ICP-like distance function
(also called ADD-S) which is robust against symmetry effects. For details, we refer to \citep{xiang2017posecnn}.
We additionally report the same metric when the translation is not applied, referred to as ``rotation only''.

\subsection{Implementation}
\label{implementation}
We implemented our experiments using the PyTorch framework \citep{paszke2017automatic},
with the Hough voting layer implemented on CPU using Numba \citep{numba}, which
proved to be more performant than a GPU implementation.
Note that there is no backpropagation through the Hough layer.

For the parts that are equivalent to PoseCNN we followed the published code,
which has some differences to the corresponding publication \citep{xiang2017posecnn},
including the application of dropout and estimation of $\log(z)$ instead of $z$ in the translation branch.
We found that these design choices improve the results in our architecture as well.

\subsection{Training}

For training ConvPoseCNN we generally follow the same approach as for PoseCNN: We use SGD with learning rate 0.001 and momentum 0.9.
For the overall loss we use $\alpha_{\text{seg}} = \alpha_{\text{trans}} = 1$. For the L2 and the Qloss we use also $\alpha_{\text{rot}} = 1$, for the SMLoss we used $\alpha_{\text{rot}} = 100$. To bring the depth error to a similar range as the center direction error, we scale the (metric) depth by a factor of 100.

We trained our network with a batch size of 2 for approximately 300,000 iterations utilizing the early stopping technique.
Since the YCB-Video Dataset contains real and synthetic frames, we choose a synthetic image
with a probability of 80\% and render it onto a random background image from the
SUN2012 \citep{xiao2010sun} and ObjectNet3D \citep{xiang2016objectnet3d} dataset.

\subsection{Prediction Averaging}
\label{sec:eval:averaging}

\begin{table}
\centering
\begin{threeparttable}
\caption{Weighting strategies for ConvPoseCNN L2}
\label{tab:resultspixell2}
\small\setlength{\tabcolsep}{.19cm}
\begin{tabular}{lcccc}
\toprule
         Method           & \multicolumn{2}{c}{ 6D pose \citep{xiang2017posecnn} }    & \multicolumn{2}{c}{Rotation only}  \\
\cmidrule(lr){2-3}\cmidrule(lr){4-5}
                      & AUC P          & AUC S          & AUC P           & AUC S      \\ \midrule
                PoseCNN \tnote{1}                      & 53.71          & 76.12          & \textbf{78.87}         & \textbf{93.16} \\ \midrule
unit weights                      & 56.59 & 78.86 & 72.87 & 90.68 \\
norm weights             & \textbf{57.13} & \textbf{79.01} & 73.84 & 91.02 \\
segm. weights & 56.63 & 78.87 & 72.95 & 90.71 \\
     \bottomrule
\end{tabular}
\begin{tablenotes}\footnotesize
 \item [1] Calculated from the PoseCNN model published in the YCB-Video Toolbox.
\end{tablenotes}
\end{threeparttable}
\end{table}

We first evaluated the different orientation loss functions presented in \cref{sec:losses}:
L2, Qloss, and SMLoss. For SMLoss, we first averaged the quaternions predicted for each object with a naive average before calculating the loss.

The next pipeline stage after predicting dense orientation is the aggregation into
a single orientation. We first investigated the quaternion average following \citep{markley2007quaternion},
using either segmentation confidence or quaternion norm as sample weights.
As can be seen in \cref{tab:resultspixell2}, norm weighting showed the best results.

\begin{table}
\centering
\begin{threeparttable}
\caption{Quaternion pruning for ConvPoseCNN L2}
\label{tab:resultsl2pruned}\small
\begin{tabular}{lcccc}
\toprule
         Method           & \multicolumn{2}{c}{ 6D pose \citep{xiang2017posecnn} }    & \multicolumn{2}{c}{Rotation only}  \\
\cmidrule(lr){2-3}\cmidrule(lr){4-5}
                      & AUC P          & AUC S          & AUC P           & AUC S      \\
\midrule
                PoseCNN                      & 53.71          & 76.12          & \textbf{78.87}         & \textbf{93.16} \\ \midrule
pruned(0)      & 57.13          & 79.01          & 73.84 & 91.02 \\
pruned(0.5)    & \textbf{57.43} & 79.14          & 74.43 & 91.33 \\
pruned(0.75)   & \textbf{57.43} & 79.19          & 74.48 & 91.45 \\
pruned(0.9)    & 57.37          & \textbf{79.23} & 74.41 & 91.50 \\
pruned(0.95)   & 57.39          & 79.21          & 74.45 & 91.50 \\
single         & 57.11          & 79.22          & 74.00 & 91.46 \\ \bottomrule
\end{tabular}
\end{threeparttable}
\end{table}

Since weighting seemed to be beneficial, which suggests that there are less
precise or outlier predictions that should be ignored, we experimented with
pruning of the predictions using the following strategy:
The quaternions are sorted by confidence and the least confident ones, according
to a removal fraction $0\leq\lambda\leq1$ are discarded.
The weighted average of the remaining quaternions is then computed as described above.
The results are shown as pruned($\lambda$) in \cref{tab:resultsl2pruned}.
We also report the extreme case, where only the most confident quaternion is left.
Overall, pruning shows a small improvement, with the ideal value of $\lambda$
depending on the target application.
More detailed evaluation shows that especially the symmetric objects show a clear
improvement when pruning. We attribute this to the fact that the averaging
methods do not handle symmetries, i.e. an average of two shape-equivalent
orientations can be non-equivalent. Pruning might help to reduce other
shape-equivalent but L2-distant predictions and thus improves the final prediction.

\subsection{Prediction Clustering}

\begin{table}
\centering
\begin{threeparttable}
\caption{Clustering strategies for ConvPoseCNN L2}
\label{tab:resultsransacl2}
\small\setlength{\tabcolsep}{.15cm}
\begin{tabular}{lcccc}
\toprule
         Method           & \multicolumn{2}{c}{6D pose \citep{xiang2017posecnn}}    & \multicolumn{2}{c}{Rotation only}  \\
\cmidrule(lr){2-3}\cmidrule(lr){4-5}
                      & AUC P          & AUC S          & AUC P           & AUC S      \\ \midrule
                PoseCNN                      & 53.71          & 76.12          & \textbf{78.87}         & \textbf{93.16} \\ \midrule
RANSAC(0.1)          & 57.18          & 79.16          & 74.12          & 91.37 \\
RANSAC(0.2)          & 57.36          & 79.20          & 74.40          & 91.45 \\
RANSAC(0.3)          & 57.27          & 79.20          & 74.13          & 91.35 \\
RANSAC(0.4)          & 57.00          & 79.13          & 73.55          & 91.14 \\
W-RANSAC(0.1) & 57.27          & 79.20          & 74.29          & 91.45 \\
W-RANSAC(0.2) & 57.42          & \textbf{79.26} & 74.53 & 91.56 \\
W-RANSAC(0.3) & 57.38          & 79.24          & 74.36          & 91.46 \\ \midrule
pruned(0.75)         & \textbf{57.43} & 79.19          & 74.48          & 91.45 \\
most confident    &   57.11	& 79.22	& 74.00	& 91.46
  \\ \bottomrule
\end{tabular}\footnotesize
RANSAC uses unit weights, while W-RANSAC is weighted by quaternion norm.
PoseCNN and the best performing averaging methods are shown for comparison.
Numbers in parentheses describe the clustering threshold in radians.
\end{threeparttable}
\end{table}

\begin{table*}
\centering
\begin{threeparttable}
\caption{6D pose, translation, rotation, and segmentation results}
\label{tab:transresults}\small\setlength{\tabcolsep}{.18cm}
\begin{tabular}{llcccccccc}
\toprule
                & & \multicolumn{2}{c}{ 6D pose \citep{xiang2017posecnn} }    & \multicolumn{2}{c}{Rotation only} &   NonSymC & SymC & Translation & Segmentation \\
\cmidrule(lr){3-4}\cmidrule(lr){5-6}\cmidrule(lr){7-7}\cmidrule(lr){8-8}\cmidrule(lr){9-9}\cmidrule(lr){10-10}
                &      & AUC P          & AUC S          & AUC P           & AUC S & AUC P           & AUC S  & Error [m] & IoU \\ \midrule

\parbox[t]{2mm}{\multirow{5}{*}{\rotatebox[origin=c]{90}{full network}}}

& PoseCNN                   & 53.71     & 76.12     & 78.87 & 93.16 & 60.49 &	63.28 & 0.0520  & 0.8369 \\
& PoseCNN (own impl.) & 53.29 &	78.31
   & 69.00 &	90.49
 & 60.91 &	57.91
 & 0.0465 & 0.8071 \\
& ConvPoseCNN Qloss         & 57.16     & 77.08     & 80.51 & 93.35 & 64.75 & 53.95
& 0.0565 & 0.7725 \\
& ConvPoseCNN Shape         & 55.54     & 79.27     & 72.15 & 91.55 & 62.77 & 56.42
 & 0.0455 & 0.8038 \\
& ConvPoseCNN L2            & 57.42     & 79.26     & 74.53 & 91.56 & 63.48 & 58.85
 & 0.0411 & 0.8044 \\
\midrule
\parbox[t]{2mm}{\multirow{4}{*}{\rotatebox[origin=c]{90}{GT segm.}}}
& PoseCNN (own impl.) & 52.90     & 80.11     & 69.60 & 91.63 & 76.63 & 84.15 & 0.0345 & 1      \\
& ConvPoseCNN Qloss         & 57.73     & 79.04     & 81.20 & 94.52 & 88.27 & 90.14 & 0.0386 & 1      \\
& ConvPoseCNN Shape         & 56.27     & 81.27     & 72.53 & 92.27 &  77.32  & 89.06 & 0.0316 & 1      \\
& ConvPoseCNN L2            & 59.50     & 81.54     & 76.37 & 92.32 & 80.67 & 85.52 & 0.0314 & 1      \\ \bottomrule
\end{tabular}\footnotesize
The average translation error, the segmentation IoU and the AUC metrics for different models. The AUC results were achieved using weighted RANSAC(0.1) for ConvPoseCNN Qloss, Markley's norm weighted average for ConvPoseCNN Shape and weighted RANSAC(0.2) for ConvPoseCNN L2.
\textit{GT segm.} refers to ground truth segmentation (i.e. only pose estimation).
\end{threeparttable}
\end{table*}

For clustering with the RANSAC strategies, we used the angular distance between
rotations as the clustering distance function and performed 50 RANSAC iterations. In contrast to the L2 distance in
quaternion space, this distance function does not suffer from the antipodal
symmetry of the quaternion orientation representation.
The results for ConvPoseCNN L2 are shown in Table \ref{tab:resultsransacl2}.
For comparison the best-performing averaging strategies are also listed.
The weighted RANSAC variant performs generally a bit better than the non-weighted variant for the same inlier thresholds, which correlates to our findings in \cref{sec:eval:averaging}.
In comparison, clustering performs slightly worse than the averaging strategies
for AUC P, but slightly better for AUC S---as expected due to the symmetry
effects.

\subsection{Loss Variants}
\begin{figure*}
\begin{tikzpicture}[font=\footnotesize]
\node[inner sep=0] (img) {\includegraphics[width=.97\textwidth]{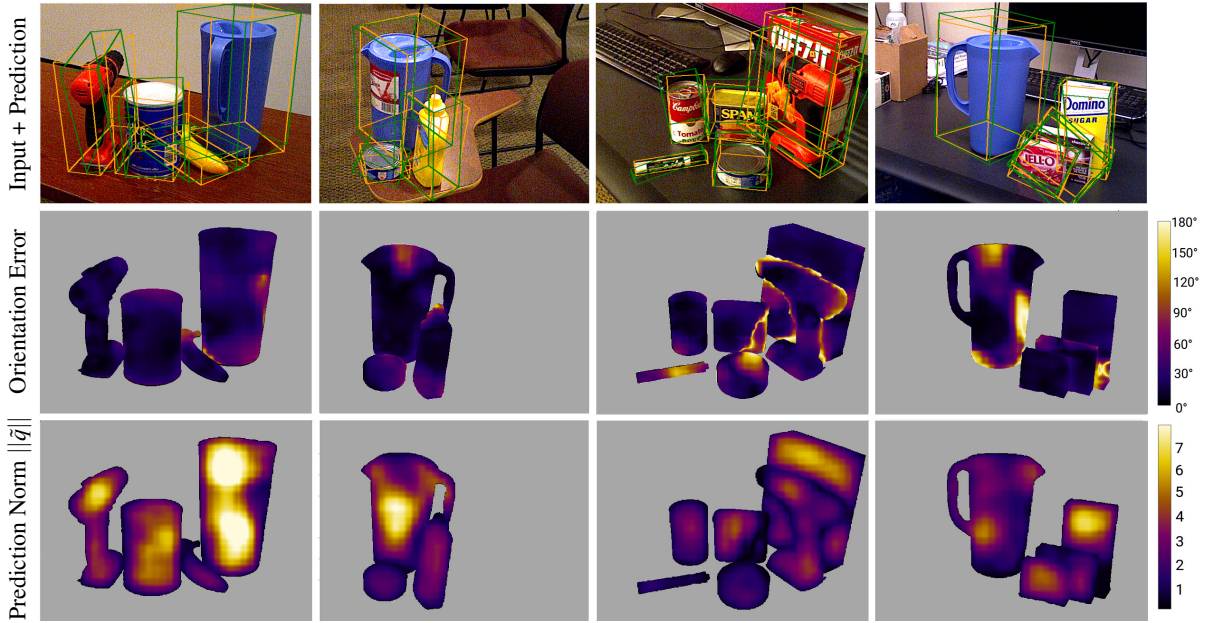}};

\node[anchor=south, rotate=90, text depth=0.25ex] at ($(img.north west)!0.16666!(img.south west)$) {Input + Prediction};
\node[anchor=south, rotate=90, text depth=0.25ex] at ($(img.north west)!0.5!(img.south west)$) {Orientation Error};
\node[anchor=south, rotate=90, text depth=0.25ex] at ($(img.north west)!0.83333!(img.south west)$) {Prediction Norm $||\tilde{q}||$};
\end{tikzpicture}
\caption{Qualitative results from ConvPoseCNN L2 on the YCB-Video test set.
Top: The orange boxes show the ground truth bounding boxes, the green boxes the 6D pose prediction.
Middle: Angular error of the dense quaternion prediction $\tilde{q}$ w.r.t. ground truth, masked by ground truth segmentation.
Bottom: Quaternion prediction norm $||\tilde{q}||$ before normalization. This measure is used for weighted aggregation.
Note that the prediction norm is low in high-error regions and high in
regions that are far from occlusions and feature-rich.}
\label{fig:qualitative_results}
\end{figure*}

\begin{table}
\centering
\begin{threeparttable}
\caption{Results for ConvPoseCNN Shape}
\label{tab:convposecnnshaperesults}\small\setlength{\tabcolsep}{.14cm}
\begin{tabular}{lcccc}
\toprule
        & \multicolumn{2}{c}{ \citep{xiang2017posecnn} Total}    & \multicolumn{2}{c}{Rotation only} \\
\cmidrule(lr){2-3}\cmidrule(lr){4-5}
                      & AUC P          & AUC S          & AUC P           & AUC S    \\ \midrule
PoseCNN  & 53.71 & 76.12     & \textbf{78.87} & \textbf{93.16}      \\ \midrule
average & 54.27   & 78.94  & 70.02 &	90.91

          \\
norm weighted    & \textbf{55.54} & 79.27  &  72.15 &	91.55  \\
pruned(0.5)                & 55.33          & \textbf{79.29}  & 71.82 &	91.45 \\
pruned(0.75)     & 54.62          & 79.09 & 70.56	 & 91.00  \\
pruned(0.85)   & 53.86          & 78.85 & 69.34	& 90.57
 \\
pruned(0.9)      & 53.23          & 78.66   & 68.37 & 90.25
   \\ \midrule
RANSAC(0.2)                  & 49.44          & 77.65   & 63.09 & 	88.73   \\
RANSAC(0.3)                  & 50.47          & 77.92       & 64.53	& 89.18      \\
RANSAC(0.4)                  & 51.19          & 78.09      & 65.61	& 89.50 \\
W-RANSAC(0.2) & 49.56  & 77.73  & 63.33 &	88.85 \\
W-RANSAC(0.3)  & 50.54   & 77.91 & 64.78 & 	89.21 \\
W-RANSAC(0.4) & 51.33  & 78.13  & 65.94	& 89.56 \\
\bottomrule
\end{tabular}
\label{tab:resultsshapeconvposecnn}
\end{threeparttable}
\end{table}
The aggregation methods showed very similar results for the Qloss trained model,
which are omitted here for brevity.
For the SMLoss variant, we report the results in Table \ref{tab:convposecnnshaperesults}. Norm weighting improves the result, but pruning does not.
This suggests that there are less-confident but important predictions with higher
distance from the mean, so that their removal significantly affects the average. This could
be an effect of training with the average quaternion, where such behavior is not
discouraged.
The RANSAC clustering methods generally produce worse results than the averaging methods in this case.
We conclude that the average-before-loss scheme is not advantageous and a fast
dense version of SMLoss would need to be found in order to apply it in our
architecture. The pixel-wise losses obtain superior performance.

\subsection{Final Results}
\label{subsec:translationresults}

\Cref{fig:qualitative_results} shows qualitative results of our best-performing
model on the YCB-Video dataset.
We especially note the spatial structure of our novel dense orientation
estimation. Due to the dense nature, its output is strongly correlated to
image location, which allows straightforward visualization and analysis of the
prediction error w.r.t. the involved object shapes.
As expected, regions that are close to boundaries between objects or far away
from orientation-defining features tend to have higher prediction error.
However, this is nicely compensated by our weighting scheme, as the predicted
quaternion norm $||\tilde{q}||$ before normalization correlates with this effect,
i.e. is lower in these regions. We hypothesize that this is an implicit effect
of the dense loss function:
In areas with high certainty (i.e. easy to recognize), the network output
is encouraged strongly in one direction.
In areas with low certainty (i.e. easy to confuse), the network cannot sufficiently
discriminate and gets pulled into several directions, resulting in outputs close to zero.

In \cref{tab:transresults}, we report evaluation metrics for our models with
the best averaging or clustering method. As a baseline, we include the PoseCNN
results, computed from the YCB-Video Toolbox model\footnote{\label{note1}\url{https://github.com/yuxng/YCB_Video_toolbox}}. We also include our re-implementation of PoseCNN.
We achieved similar final AUCs on the test set.
We also show more detailed results with regard to translation and segmentation of the different models.
For this we report the average translation error and the segmentation IoU for all models in Table \ref{tab:transresults}. They show that there is a strong influence of the translation estimation on the AUC losses. However, for the models with better translation estimation, the orientation estimation is worse. 

For the total as reported by PoseCNN, all three Conv\-Pose\-CNNs have a bit higher AUC than PoseCNN, but only the model trained with Qloss has a similar orientation estimation to PoseCNN.
Compared to PoseCNN, some models perform better for the orientation and some better for the translation even though the translation estimation branch is the same for all of these networks. We were interested in the models performance with regard to the symmetric and non-symmetric objects. For this we calculated the class-wise average over the AUCs for the symmetric and non-symmetric objects separately. In Table \ref{tab:transresults} we report them as NonSymC and SymC and report AUC P and AUC S respectively. ConvPoseCNN performed a bit better than PoseCNN for the non-symmetric objects but worse for the symmetric ones. 
This is not surprising since Qloss and L2 loss are not designed to handle symmetric objects. 
The model trained with SMLoss also achieves suboptimal results for the symmetric objects compared to PoseCNN. This might be due to different reasons: First, we utilize an average before calculating the loss; therefore
during training the average might penalize predicting different shape-equivalent quaternions, in case their average is not shape-equivalent. Secondly, there are only five symmetric objects in the dataset and we noticed that two of those, the two clamp objects, are very similar and thus challenging, not only for the orientation but as well for the segmentation and vertex prediction. This is further complicated by a difference in object coordinate systems for these two objects.

We also included results in Table \ref{tab:transresults} that were produced by evaluating using the ground truth semantic segmentation, in order to investigate how much our model's performance could improve by the segmentation performance alone. If the segmentation is perfect, then the orientation and the translation estimation of all models improve. 
Even the re-implemented PoseCNN improves its orientation; therefore the RoIs must have improved by the better translation and inlier estimation.
Even though our aim was to change the orientation estimation of PoseCNN, our results show that this cannot be easily isolated from the translation estimation, since both have large effects on the resulting performance.
In our experiments, further re-balancing of the loss coefficients was not productive
due to this coupled nature of the translation and orientation sub-problems.

We conclude that finding a proper balancing between translation and orientation estimation is important but difficult to achieve. Also, a better segmentation would further improve the results.

\section{Comparison to Related Work}

\begin{table}
\centering
\begin{threeparttable}
\caption{Comparison to Related Work}
\label{tab:rwcomparison}\small
\begin{tabular}{lccc}
\toprule
                                     & \multicolumn{2}{c}{Total} &  Average \\
\cmidrule(lr){2-3}\cmidrule(lr){4-4}
                                     & AUC P       & AUC S     & AUC \citep{heatmaps}    \\ \midrule
PoseCNN                      & 53.7        & 75.9        & 61.30  \\
ConvPoseCNN L2 & 57.4        & 79.2         & 62.40  \\
\citep{heatmaps} without FM      &            &                    & 61.41             \\ \midrule
ConvPoseCNN+FM  & 58.22 & 79.55 & 61.59 \\
\citep{heatmaps} with FM       &            &                    & 72.79     \\ \bottomrule
\end{tabular}\footnotesize
Comparison between PoseCNN (as reported in \citep{xiang2017posecnn}), ConvPoseCNN L2 with pruned(0.75), and \citet{heatmaps} without and with Feature Mapping (FM).
\end{threeparttable}
\end{table}

\begin{table}
\caption{Detailed Class-wise Results}\label{tab:bestconvposecnn}
\footnotesize   \setlength{\tabcolsep}{.15cm}
\begin{tabular}{lrrrrr} \toprule
Class                                                  & \multicolumn{2}{c}{Ours} & \multicolumn{2}{c}{PoseCNN} \\
\cmidrule(lr){2-3}\cmidrule(lr){4-5}
                                                       & AUC P        & AUC S  &  AUC P       & AUC S \\
\midrule
master$\_$chef$\_$can                           & 62.32        & 89.55  &  50.08       & 83.72 \\
cracker$\_$box                                  & 66.69        & 83.78  &  52.94       & 76.56 \\
sugar$\_$box                                    & 67.19        & 82.51  &  68.33       & 83.95 \\
tomato$\_$soup$\_$can                           & 75.52        & 88.05  &  66.11       & 80.90 \\
mustard$\_$bottle                               & 83.79        & 92.59  &  80.84       & 90.64 \\
tuna$\_$fish$\_$can                             & 60.98        & 83.67  &  70.56       & 88.05 \\
pudding$\_$box                                  & 62.17        & 76.31  &  62.22       & 78.72 \\
gelatin$\_$box                                  & 83.84        & 92.92  &  74.86       & 85.73 \\
potted$\_$meat$\_$can                           & 65.86        & 85.92  &  59.40       & 79.51 \\
banana                                          & 37.74        & 76.30  &  72.16       & 86.24 \\
pitcher$\_$base                                 & 62.19        & 84.63  &  53.11       & 78.08 \\
bleach$\_$cleanser                              & 55.14        & 76.92  &  50.22       & 72.81 \\
bowl                    & 3.55         & 66.41  &  3.09        & 70.31 \\
mug                                             & 45.83        & 72.05  &  58.39       & 78.22 \\
power$\_$drill                                  & 76.47        & 88.26  &  55.21       & 72.91 \\
wood$\_$block           & 0.12         & 25.90  &  26.19       & 62.43 \\
scissors                                        & 56.42        & 79.01  &  35.27       & 57.48 \\
large$\_$marker                                 & 55.26        & 70.19  &  58.11       & 70.98 \\
large$\_$clamp          & 29.73        & 58.21  &  24.47       & 51.05 \\
extra$\_$large$\_$clamp & 21.99        & 54.43  &  15.97       & 46.15 \\
foam$\_$brick           & 51.80        & 88.02  &  39.90       & 86.46 \\
\bottomrule
\end{tabular}
\end{table}

In Table \ref{tab:rwcomparison} we compare ConvPoseCNN  L2, to the values reported in the PoseCNN paper, as well as with a different class-wise averaged total as in \citep{heatmaps}. We also compare to the method of \citet{heatmaps}, with and without their proposed Feature Mapping technique, as it should be orthogonal to our proposed method.
One can see that our method slightly outperforms PoseCNN, but we make no claim
of significance, since we observed large variations depending on various
hyperparameters and implementation details.
We also slightly outperform \citet{heatmaps} without Feature Mapping.
\Cref{tab:bestconvposecnn} shows class-wise results.

We also investigated applying the Feature Mapping technique \citep{heatmaps} to our model.
Following the process, we render synthetic images with poses corresponding to the real training data. We selected the extracted VGG-16 features for the mapping process and thus have to transfer two feature maps with 512 features each. Instead of using a fully-connected architecture as the mapping network, as done in \citep{heatmaps}, we followed a convolutional set-up and mapped the feature from the different stages to each other with residual blocks based on $(1 \times 1)$ convolutions. %

The results are reported in Table \ref{tab:rwcomparison}. However, we did not observe
the large gains reported by \citet{heatmaps} for our
architecture.
We hypothesize that the feature mapping
technique is highly dependent on the quality and
distribution of the rendered synthetic images, which are maybe not of sufficient quality in our case.

\section{Time Comparisons}
\label{sec:timecomparison}

\begin{table}
\centering
\begin{threeparttable}
\caption{Training performance \& model sizes}
\label{tab:timecomparisontrain}\small
\begin{tabular}{lcl} \toprule
                       & Iterations/s\tnote{1} & Model size \\ \midrule
PoseCNN                & 1.18          & 1.1 GiB     \\
ConvPoseCNN L2         & 2.09          & 308.9 MiB   \\
ConvPoseCNN Qloss      & 2.09          & 308.9 MiB   \\
ConvPoseCNN Shapeloss & 1.99          & 308.9 MiB   \\ \bottomrule
\end{tabular}\footnotesize
\begin{tablenotes}
 \item [1] Using a batch size of 2. Averaged over 400 iterations.
\end{tablenotes}
\end{threeparttable}
\end{table}

We timed our models on an NVIDIA GTX 1080 Ti GPU with 11\,GB of memory.
Table \ref{tab:timecomparisontrain} lists the training times for the different models, as well as the model sizes when saved. 
The training of the Conv\-Pose\-CNNs is almost twice as fast and the models are
much smaller compared to PoseCNN.

\begin{table}
\centering
\begin{threeparttable}
\caption{Inference timings}
\label{tab:testtime}\small
\begin{tabular}{lrr} \toprule
Method                    & Time [ms]\tnote{1}  & Aggregation [ms] \\ \midrule
PoseCNN \citep{xiang2017posecnn} & 141.71 &                  \\ \midrule
ConvPoseCNN & & \\
- naive average         & 136.96 & 2.34             \\
- average               & 146.70 & 12.61            \\
- weighted average      & 146.92 & 13.00            \\
- pruned w. average & 148.61 & 14.64            \\
- RANSAC                & 158.66 & 24.97 \\
- w. RANSAC       & 563.16 & 65.82         \\ \bottomrule
\end{tabular}\footnotesize
\begin{tablenotes}
 \item [1] Single frame, includes aggregation.
\end{tablenotes}
\end{threeparttable}
\end{table}

The speed of the ConvPoseCNN models at test time depends on the method used for quaternion aggregation. The times for inference are shown in \cref{tab:testtime}. %
For the averaging methods the times do not differ much from PoseCNN. PoseCNN takes longer to produce the output, but then does not need to perform any other step. For ConvPoseCNN the naive averaging method is the fastest, followed by the other averaging methods.  RANSAC is, as expected, slower. The forward pass of ConvPoseCNN takes about 65.5\,ms, the Hough transform around 68.6\,ms. We note that the same Hough transform implementation is used for PoseCNN and ConvPoseCNN in this comparison.

In summary, we gain advantages in terms of training time and model size, while
inference times are similar. While the latter finding initially surprised us, we
attribute it to the high degree of optimization that RoI pooling methods
in modern deep learning frameworks have received.

\section{Conclusion}

As shown in this work, it is possible to directly regress 6D pose parameters
in a fully-convolutional way, avoiding the sequential cutting out and normalizing of individual object hypotheses.
Doing so yields a much smaller, conceptually simpler architecture with fewer
parameters that estimates the poses of multiple objects in parallel. 
We thus confirm the corresponding trend in the related object
detection task---away from RoI-pooled architectures towards fully-convolutional ones---also for the pose estimation task.

We demonstrated benefits of the architecture in terms of the number of parameters
and training time without reducing prediction accuracy on the YCB-Video dataset.
Furthermore, the dense nature of the orientation prediction allowed us to
visualize both prediction quality and the implicitly learned weighting and thus
to confirm that the method attends to feature-rich and non-occluded regions.

An open research problem is the proper aggregation of dense
predictions. While we presented methods based on averaging and clustering,
superior (learnable) methods surely exist. In this context, the proper handling
of symmetries becomes even more important.
In our opinion, semi-supervised methods that learn object symmetries and thus
do not require explicit symmetry annotation need to be developed, which is
an exciting direction for further research.

\vspace{1ex}\textit{Acknowledgment:}
{\footnotesize
This work was funded by grant BE 2556/16-1 (Research Unit FOR 2535Anticipating Human Behavior) of the German Research Foundation (DFG).
}

\printbibliography

\end{document}